%% file: StructPR_revise_3.tex
\documentclass[journal,11pt,onecolumn,draftclsnofoot]{IEEEtran} 
\usepackage{epsfig,algorithm,amsmath,amssymb,color,subfigure,hyperref}
\usepackage{amssymb,amsmath,algorithm}
\usepackage{amsfonts,amsthm, bm, bbm,dsfont}
\usepackage{booktabs, cite} 
\usepackage{pgfplotstable}

\begin{document}

\newcommand{\bigo}{\mathcal{O}}


\newcommand{\p}{\bm{p}}
\renewcommand{\j}{\bm{j}}
\newcommand{\uu}{{\bm{u}}}
\newcommand{\tot}{\mathrm{tot}}
\renewcommand{\a}{\bm{a}}
\newcommand{\ta}{\bm{\tilde{a}}}
\newcommand{\x}{\bm{x}}
\newcommand{\e}{\bm{e}}
\newcommand{\h}{\bm{h}}
\renewcommand{\b}{\bm{b}}
\newcommand{\y}{\bm{y}}
\newcommand{\w}{\bm{w}}
\newcommand{\X}{\bm{X}}
\newcommand{\U}{{\bm{U}}}
\newcommand{\R}{{\bm{R}}}
\newcommand{\B}{\bm{B}}
\newcommand{\G}{\bm{G}}

\newcommand{\I}{\bm{I}}
\newcommand{\Y}{\bm{Y}}
\newcommand{\g}{\bm{g}}
\newcommand{\s}{\bm{s}}
\newcommand{\dd}{\bm{d}}
\newcommand{\vv}{\bm{v}}
\newcommand{\ghat}{\bm{\hat{g}}}
\newcommand{\xhat}{\bm{\hat{x}}}
\newcommand{\bhat}{\bm{\hat{b}}}
\newcommand{\Uhat}{\hat\U}
\newcommand{\uhat}{{\bm{\hat{u}}}}
\newcommand{\Zhat}{{\bm{\hat{Z}}}}
\newcommand{\Z}{{\bm{{Z}}}}
\newcommand{\D}{{\bm{D}}}
\newcommand{\Q}{{\bm{Q}}}
\newcommand{\F}{{\bm{F}}}
\newcommand{\bP}{{\bm{P}}}

\newcommand{\n}{\mathcal{N}}
\newcommand{\E}{\mathbb{E}}
\newcommand{\one}{\mathds{1}}

\newcommand{\z}{\bm{z}}
\newcommand{\indic}{ \bm{1} }

\newcommand{\gd}{\bm{v}}
\newcommand{\ghatd}{\hat{\gd}}
\renewcommand{\ng}{\nu}
\newcommand{\hatng}{\hat{\ng}}

\newcommand{\iidsim}{\stackrel{\mathrm{iid}}{\thicksim }}
\newcommand{\indepsim}{\stackrel{\mathrm{indep.}}{\thicksim }}
\newcommand{\SE}{\mathrm{SE}}
\newcommand{\dist}{\mathrm{dist}}
\newcommand{\tta}{\ta^{\text{trunc}}}

\newcommand{\V}{\bm{V}}
\newcommand{\A}{\bm{A}}
\newcommand{\C}{\bm{C}}
\newcommand{\Chat}{\bm{\hat{C}}}
\newcommand{\cb}{\bm{c}}

\newcommand{\M}{\bm{M}}

\newcommand{\Span}{\mathrm{Span}}
\newcommand{\trace}{\mathrm{trace}}
\newcommand{\rank}{\mathrm{rank}}

\newcommand{\cM}{\hat{\bm{D}}}  
\newcommand{\cA}{\bm{D}}    
\newcommand{\cH}{\bm{H}}
\newcommand{\bE}{\bm{E}}
\newcommand{\bF}{\bm{F}}
\newcommand{\total}{\mathrm{tot}}
\newcommand{\Lamk}{\bm\Lambda_k}
\newcommand{\Lambar}{\bar{\bm\Lambda}}
\newcommand{\evdeq}{\overset{\mathrm{EVD}}=} 
\newcommand{\svdeq}{\overset{\mathrm{SVD}}=} 
\newcommand{\qreq}{\overset{\mathrm{QR}}=} 
\newcommand{\lammax}{\bar\lambda_{\max}} 
\newcommand{\lammin}{\bar\lambda_{\min}}

\newcommand{\bz}{\boldsymbol{z}}

\newcommand{\N}{\mathcal{N}}
\renewcommand{\S}{\mathcal{S}}
\newcommand{\W}{\bm{W}}
\renewcommand{\Re}{\mathbb{R}}

\newcommand{\matdist}{\text{mat-dist}}
\newcommand{\Bstar}{{\B^*}}
\newcommand{\bstar}{\b^*}
\newcommand{\tB}{\B^*} 
\newcommand{\tb}{\b^*}
\newcommand{\td}{\tilde{\bm{d}}^*}
\newcommand{\init}{{\mathrm{init}}}
\newcommand{\bea}{\begin{eqnarray}} 
\newcommand{\eea}{\end{eqnarray}}

\newcommand{\Ustar}{\U^*}
\newcommand{\Xstar}{{\X^*}}
\newcommand{\xstar}{\x^*}
\newcommand{\deltinit}{\delta_\init}
\newcommand{\deltapt}{\delta_{t}}
\newcommand{\deltaptplus}{\delta_{t+1}}

\newcommand{\bSigma}{{\bm\Sigma^*}}
\newcommand{\tSigma}{\bm{E}_{det}}
\newcommand{\sigmin}{{\sigma_{\min}^*}}
\newcommand{\sigmax}{{\sigma_{\max}^*}}

\newcommand{\HH}{{\bm{D}}}
\newcommand{\GG}{{\bm{M}}}
\newcommand{\SSS}{{\bm{S}}}
\newcommand{\ik}{{ik}}
\newcommand{\full}{{\mathrm{full}}}
\newcommand{\qfull}{q_\full}
\newcommand{\sub}{{\mathrm{sub}}}

\newcommand{\T}{\mathcal{T}}
\newcommand{\That}{\hat\T}

\renewcommand{\SE}{\sin \Theta}

\newcommand{\checkU}{\U} \newcommand{\checktB}{\B} \newcommand{\checktb}{\b}

\newcommand{\beq}{\begin{equation} }
\newcommand{\eeq}{\end{equation} }
\newcommand{\bi}{\begin{itemize}} 
\newcommand{\ei}{\end{itemize}}
\newcommand{\ben}{\begin{enumerate}}
\newcommand{\een}{\end{enumerate}}

\newcommand{\cblue}{} 
\newcommand{\cbl}{} 


%

%

\title{Non-Convex Structured Phase Retrieval} 
\author{Namrata Vaswani \\
Dept of Electrical and Computer Engineering, \\
Iowa State University, Ames IA, USA
}
\maketitle

\begin{abstract}
Phase retrieval (PR), also sometimes referred to as quadratic sensing, is a problem that occurs in numerous signal and image acquisition domains ranging from optics, X-ray crystallography, Fourier ptychography, sub-diffraction imaging, and astronomy. In each of these domains, the physics of the acquisition system dictates that only the magnitude (intensity) of certain linear projections of the signal or image can be measured. Without any assumptions on the unknown signal, accurate recovery necessarily requires an over-complete set of measurements. The only way to reduce the measurements/sample complexity is to place extra assumptions on the unknown signal/image. A simple and practically valid set of assumptions is obtained by exploiting the structure inherently present in many natural signals or sequences of signals.
Two commonly used structural assumptions are (i) sparsity of a given signal/image or (ii) a low rank model on the matrix formed by a set, e.g., a time sequence, of signals/images. Both have been explored for solving the PR problem in a sample-efficient fashion. This article describes this work, with a focus on non-convex approaches that come with sample complexity guarantees under simple assumptions. We also briefly describe other different types of structural assumptions that have been used in recent literature.
\end{abstract}


\section{Introduction}
\label{intro}

Phase retrieval (PR) is a problem that occurs in numerous signal and image acquisition domains ranging from optics, X-ray crystallography, Fourier ptychography, sub-diffraction imaging, and astronomy. In each of these domains, the physics of the acquisition system dictates that only the magnitude (intensity) of certain linear projections of the signal or image can be measured. For example, often the linear projections are discrete Fourier transform coefficients of the signal or image, or of their masked versions. In all of these applications, the phase is difficult or impossible to obtain. Another important application of PR is in latent variable models where the hidden variables are the missing signs of the linear projections of the unknown data vector/signal \cite{nonconvex_review}.


Mathematically, the goal of PR is to recover an $n$-length signal $\xstar$ from measurements $\y_i:=| \langle \a_i, \xstar \rangle|$, $i=1,2, \dots, m$. The measurement vectors $\a_i$ are known. This general PR problem is often also referred to as {\em quadratic sensing}.
Algorithmic heuristics for solving (Fourier) PR have existed since the early works of Gerchberg and Saxton\cite{ger_saxton} from the 1970s.
In recent years, there has been much renewed interest in PR in the signal processing community with the goal of obtaining fast and provably correct solution solutions.
Most provable guarantees assume that the measurement vectors $\a_i$ are (real or complex) independent identically distributed (i.i.d.)  standard Gaussian vectors since this is the simplest model under which algorithms can be analyzed\footnote{Rotational symmetry of the standard Gaussian allows use of nice tricks to compute or bound expected values, and its light-tailed property allows one to use existing concentration inequalities \cite{versh_book}.}. Whenever this is assumed we refer to the problem as {\em ``standard''} PR. A few existing guarantees also hold for the coded diffraction pattern (masked Fourier) setting.

Without any assumptions on the signal $\xstar$, accurate recovery necessarily requires an over-complete set of measurements, i.e., $m$ needs to be equal to or larger than  $n$. This requirement can be a challenge when moving to very high resolution imaging~\cite{holloway} because it implies a proportionally higher cost of data acquisition (in terms of time, number of sensors, or power consumption). Low-cost dynamic imaging of scenes exhibiting a temporal evolution, e.g., imaging of live biological samples, poses an even greater challenge: in order to  be able to capture changes in the scene, the image acquisition process needs to be fast enough.
For example, Fourier ptychography is a technique for super-resolution in which each of a set of low resolution cameras measures the magnitude of a different band-pass filtered version of the target high-resolution image. To get enough measurements per image, one either needs many cameras (expensive), or one needs to move a single camera to different locations to acquire the different bands (slow) \cite{holloway,TCIgauri}. The latter is a low-cost  option, but it makes the acquisition process very slow.
The question is can we use practically valid assumptions on the signal(s) that can enable high-resolution image reconstruction using fewer total measurements (in this example, fewer cameras or fewer ``on'' pixels per camera)?

A simple way to do this is to exploit the structure inherently present in many natural signals or sequences of signals.
Two commonly used structural assumptions are (i) sparsity of a given signal/image or (ii) a low rank model on the matrix formed by a set, e.g., a time sequence, of signals/images. Both of these have been extensively used to speed up imaging in many biomedical imaging applications in which image acquisition is a slow process. An important example is Compressive Sensing (CS) for Magnetic Resonance Imaging (MRI) or more generally Compressive MRI.
The low-rank model is also an important component of many practically useful approaches to Compressive dynamic MRI, e.g., see \cite{st_imaging,dyn_mri1,lee2017near}. 

Both sparsity and low-rank have been explored for solving PR (quadratic sensing) problems in a sample-efficient fashion. 
This article describes this work, with a focus on  {\em non-convex approaches} that {\em come with sample complexity guarantees (required lower bounds on the number of measurements/samples $m$)}. We also briefly describe other different types of structural assumptions that have been used in recent literature -- dynamic sparsity or low-rank, deep neural network based prior models, and compression priors -- and the pros and cons of using the different assumptions. We begin the article with explaining non-convex approaches, followed by a brief review of unstructured PR literature, and then a short discussion of the tools used in the theoretical analysis.  We end with a discussion of interesting open questions.


\section{Non-convex approaches}
\cblue A signal recovery problem is called {\em non-convex} if the optimization problem to be solved is non-convex\footnote{The cost function is not convex or the feasible set of the constraints is not convex or both.} and it cannot be reformulated to get a convex program without relaxing any assumptions. 
Some common examples include Compressive Sensing (CS) / sparse recovery, low-rank matrix completion (LRMC) and matrix sensing (LRMS).

We define these problems here because we frequently refer to them later. CS involves recovering an $s$-sparse $n$-length vector $\xstar$ from $\y:=\A \xstar$ when $\A$ is an $m \times n$ matrix with $m < n$ (under-determined system). When $\A$ has i.i.d. standard Gaussian entries, we refer to this problem as standard CS.
LRMS involves recovering a rank-$r$ $n \times q$ matrix $\Xstar$ from measurements $\y_i:=\langle \A_i, \Xstar \rangle  , i=1,2, \dots, m$ where $\A_i$ are dense (non-sparse) matrices. Standard LRMS again means that the $\A_i$'s are i.i.d. and each contains i.i.d. standard Gaussian entries. LRMC on the other hand involves recovering $\Xstar$ from a subset of its entries, thus, it is LRMS with $\A_i$'s being one-sparse matrices. Standard LRMC assumes a Bernoulli($\rho$) model on the set of observed entries: a matrix entry is observed with probability $\rho$ independent of all others.
In case of CS and LRMS, each scalar measurement $\y_i$ is a function of the entire unknown signal or matrix. This type of measurements are referred to as {\em global}.  This is not the case for LRMC though. 

\cbl
For all these non-convex problems, one can define a ``relaxed'' problem that is convex and prove that, under certain assumptions, the solution of the convex problem is unique and equal to that of the original one. For CS, this is typically done via $\ell_1$ minimization, while for LRMC and LRMS, the most common convex relaxation is nuclear norm minimization. \cblue
An alternative solution approach is to come up with iterative algorithms to directly solve the original non-convex problem.  This category includes alternating minimization (AltMin) algorithms, such as AltMinComplete/AltMinSense for LRMC/LRMS, projected gradient descent (GD) methods such as iterative hard thresholding (IHT) for CS or for LRMS and LRMC, or greedy solutions such as Orthogonal Matching Pursuit for CS.
\cbl
{\em All such iterative algorithms for directly solving a non-convex problem are commonly referred to as  ``non-convex approaches''}. We will use this term often.

\cblue
Projected GD is a simple modification of the GD idea for constrained optimization. At each iteration, after one GD step, the algorithm projects the new estimate onto the constraint set (finds the entry in the constraint set that is closest to it in the chosen distance metric, usually $l_2$ norm).  Thus, for example, if the constraint set is the set of $s$-sparse vectors, then the projection step involves zeroing out all but the $s$ largest magnitude entries of the vector. If the set is rank $r$ matrices, then, it involves computing the $r$-SVD of the matrix.
AltMin splits the unknown variables into two parts, and at each iteration, minimizes over one variable keeping the other fixed at its previous estimate, followed by vice versa. The variables are chosen so that each of the two minimizations is ``easy'' (either closed form or a known approach exists).
%
%
While both AltMin and projected GD have been in use for very long, provable guarantees for these have started appearing only in the last decade, see \cite{optspace,lowrank_altmin}, and follow-up works. The key ingredient of most of these algorithms is a carefully designed spectral initialization step that provides an initial estimate that is close enough to the true signal.
 The idea is come up with a computable matrix that is such that (i) its expected value has the ``correct'' top left singular vector(s), and  (ii) it is close to its expected value with high probability (w.h.p.).
Here ``correct'' means the following. When recovering a signal $\xstar$, we want the top singular vector of the expected value matrix to be proportional to $\xstar$. When recovering a rank-$r$ matrix $\Xstar$, we want the top $r$ left singular vectors to span the column span of $\Xstar$. 


\cblue
Non-convex algorithms (with their carefully designed initialization) are interesting because they are typically much faster than solvers for the convex relaxation, both in terms of theoretical complexity and in practice. For solvers for many convex programs, either the per iteration complexity is significantly more than linear in the problem size or the number of iterations required grows as $1/\sqrt{\epsilon}$ or more, or both \cite{lowrank_altmin,robpca_nonconvex}. Here $\epsilon$ is the desired accuracy: relative error between the solution produced by the solver and the true (unknown) solution of the convex program.
{\em An exception is $l_1$ minimization or basis pursuit for which nearly linear complexity solvers now exist.}
On the other hand, for non-convex methods, the per iteration complexity is typically linear or close to linear\footnote{``close to linear'' means linear times log factors or linear times a polynomial in $r$ or in $s$ where $r$ ($s$) is the small rank (support size).}, and one can often prove geometric convergence (when starting from the carefully designed initialization) so that the number of iterations required is proportional to $\log(1/\epsilon)$ \cite{lowrank_altmin,robpca_nonconvex}.
\cbl

\section{Brief review of standard unstructured PR / Quadratic Sensing}
We provide a brief review of the literature on non-convex approaches to {\em unstructured}  PR in order to make it easier to understand the structured PR approaches. 
By arranging the measurement vectors $\a_i$ as the rows of an $m \times n$ matrix $\A$, i.e., by letting  $\A:=[\a_1, \a_2,\dots \a_m]'$, the PR problem can be rewritten as: recover $\xstar$ from
\beq
\y:= |\A \xstar|
\label{y_mod}
\eeq
{\em Here and in the rest of the paper, $|.|$ is applied element-wise to each entry of the vector, and $'$ denotes matrix or vector (conjugate) transpose.}  
We focus on {\em standard PR}, i.e., the $\a_i$'s are (real or complex) i.i.d. standard Gaussians.
\cblue We should mention that the real-valued Gaussian measurements' case is not directly a special case of the complex-valued case. But, it can converted into a special case if we use the squared sum of two real measurements as ``one'' complex measurement squared. \cbl



Clearly, with the PR measurement model, one cannot distinguish $\x$ from $-\x$. Similarly, when $\x$ is a complex-valued vector, $\x$ and $\x e^{j \theta}$, for any angle $\theta \in [0,2\pi]$, generate the exact same set of measurements and, thus, cannot be distinguished. Hence a meaningful metric for success for PR  is the phase (or sign) invariant distance defined as
\[
\dist(\xhat,\x):= \min_{\theta  \in [0,2\pi]} \| \x e^{j \theta} - \xhat\|_2^2.
\]
When $\x$ is real-valued, this simplifies to $\dist(\xhat,\x) = \min(\|\xhat-\x\|, \|\xhat+ \x\|)$.
Thus, a PR solution $\xhat$ is an {\em ``$\epsilon$-accurate''} of $\x$ if $\dist^2(\xhat,\x)/\|\x \|_2^2 \le \epsilon^2$. For recovering an $n \times q$ matrix $\X$, it means that $\sum_{k=1}^q \dist^2(\xhat_k,\x_k)/\|\Xstar \|_F^2 \le \epsilon^2$.



The first work on provable standard PR \cite{candes_phaselift} consisted of solving a convex relaxation to recover the rank-one matrix $\bm{Z}:=\x \x'$, given its \emph{linear} measurements $\y_i^2:= \langle \a_i, \x \rangle^2 =  \langle \a_i \a_i{}', \bm{Z} \rangle$. The vector $\x$ was then estimated as the top eigenvector of the recovered matrix. This approach, called \emph{PhaseLift}, could provably recover $\x$ (up to a global phase uncertainty) using only $m \in \Omega( n)$ i.i.d. Gaussian measurements\footnote{$f(n) \in \Omega(g(n))$ means that there exists a constant $C >0$ and an integer $n_0 < \infty$, such that for all $n \ge n_0$, $f(n) \ge C g(n)$. $f(n) \in \bigo(g(n))$ means that there exists a constant $C >0$ and an integer $n_0 < \infty$, such that, for all $n \ge n_0$, $f(n) \le C g(n)$.}.
However, due to the ``lifting" to an $n^2$ dimensional problem, the resultant algorithm was both very slow and space inefficient.
%
In recent work, faster  \emph{non-convex methods}, that do not lift to higher dimensions and are much faster, have been explored~\cite{pr_altmin,altmin_irene_w,wf,twf}.  The first such piece of work studied the classical Gerchberg-Saxton algorithm with two modifications: (i) a novel spectral initialization was used, and (ii) the algorithm used ``sample-splitting'' (a different independent set of $m$ measurements was used in each new iteration) \cite{pr_altmin}. The authors termed this approach AltMinPhase. 
It computes the initial estimate $\xhat^{(0)}$ as the top singular vector of
\beq
\Y_0 := \frac{1}{m} \sum_{i=1}^m \y_i^2 \a_i \a_i{}'
\label{def_Y0}
\eeq
One can show that $\E[\Y_0] = \|\xstar\|^2 \I + 2 \xstar \xstar{}'$ and thus its top singular vector is proportional to $\xstar$ \cite{pr_altmin}.
Using an appropriate concentration bound \cite{versh_book}, along with the Davis-Kahan $\sin \Theta$ theorem, one can then also show that the top singular vector of $\Y_0$ will be close to $\xstar$ with high probability (w.h.p.). With this initialization, AltMinPhase alternates between the following two steps: at each iteration $t$,
\ben
\item estimate the diagonal matrix $\C$ of measurements' phases, $\hat\C_{i,i}^{(t)}:= \mathrm{phase}(\langle \a_i, \xhat^{(t-1)} \rangle)$, $i=1,2,\dots,m$; here $\mathrm{phase}(z):= z/|z|$.
\item use this to obtain a new signal estimate by solving the following least squares (LS) problem:\\
$\xhat^{(t)} := \arg \min_{\x} \|\y - \hat\C' \A \x\|_2^2$ with $\hat\C = \hat\C^{(t)}$.
\een
It was shown that, if  $m \in \Omega( n \log^3 n \log (1/\epsilon) )$, then, w.h.p., AltMinPhase converges to an $\epsilon$ accurate solution of standard PR in $\bigo(\log(1/\epsilon))$ iterations.

In later work, a GD method with the same spectral initialization as above was studied.  This method, named Wirtinger Flow (WF), implemented GD to minimize
\beq
\frac{1}{m} \sum_{i=1}^m (\y_i^2 - |\langle \a_i, \x \rangle|^2 )^2.
\label{costfn}
\eeq
WF needed only $\Omega( n \log n )$ measurements and did not use sample-splitting. But the guarantee required the GD step size to be proportional to $1/n$ and hence the number of iterations required was proportional to $n$, making its time complexity $\bigo(mn^2)$.
In \cite{twf}, a truncation idea was introduced in both the spectral initialization and the gradient steps of WF; the resulting method was called Truncated WF (TWF). This simple, but important, modification helped ensure that TWF could recover $\xstar$ using only $m \in \Omega (n)$ independent Gaussian phaseless measurements {\em and} with constant step size so that its computational complexity was only $\bigo(mn \log (1/\epsilon))$. 
This approach was order-optimal in sample complexity and nearly linear in computational complexity as well.
%
\cblue In follow-up work, an order-optimal sample complexity guarantee was obtained for the AltMin solution as well when used in conjunction with the truncated spectral initialization \cite{altmin_irene_w}.
\cbl
Two other follow-up works developed Reshaped WF (RWF) and Truncated Amplitude Flow (TAF) that also had similar properties.
%
Finally, newer work shows that GD for PR can succeed even with random initialization, but requires more than order $n$ measurements in this case, see the references in \cite{nonconvex_review}.

\cblue
A different line of work studies  convex relaxation approaches such as PhaseMax that do not lift the PR problem to higher dimensions.
Unlike PhaseLift though, these assume that a good approximation to the unknown signal is available. When used with the truncated spectral initialization of \cite{twf}, PhaseMax also achieves exact recovery with $m \ge C n$. Moreover the PhaseMax convex program can be solved using $l_1$ minimization making it about as fast as the non-convex approaches described above.
\cbl




\cblue
\renewcommand{\O}{\bm{O}}

\section{Proof Ideas}
Proofs for PR or structured PR typically rely on the use of the following tools: (i) the Davis-Kahan $\sin \Theta$ theorem, (ii) the definition of a matrix  $l_2$ norm as $\|\M\|_2 = \max_{\x:\|\x\|_2 = 1} \|M \x\|_2$ and its vector version as $\|\z\|:=  \max_{\x: \|\x\|_2 = 1} |\z' \x|$, (iii) the ``$\epsilon$-net argument'' for covering the surface of a unit hyper-sphere by a finite number of balls of radius $\epsilon$, and (iv) concentration bounds for sums of independent random variables or matrices. In particular, the matrix Bernstein inequality and the sub-exponential Bernstein inequality (or results that rely on it) are used often.  All these tools are nicely explained in \cite{versh_book}.
Moreover, the Cauchy-Schwarz inequality is frequently used, both the expected value version and the simpler version for sums of products of deterministic scalars or matrices. Of course, its bounds can be loose, and can result in sub-optimal sample complexity bounds. These have often been improved in later works by finding a way to either eliminate its use or to postpone its use until a later point in the proof.  For an example, see \cite{lrpr_improved}. 

To compute or bound terms of the form $\E[ f ( \a'\z, \a  )]$, when $\a$ is a standard Gaussian vector that is independent of the vector $\z$, the following idea is commonly used to simplify computations. Here $\E[.]$ denote expectation over the distribution of $\a$.
Carefully pick a {\em unitary} matrix $\O$ that is independent of $\a$ but can depend on $\z$. Since $\a$ is standard Gaussian, $\tilde\a:=\O'\a$ is also standard Gaussian.
Thus, $\E[ f ( \a'\z, \a  )] =  \E[ f(\a'\O \O' \z, \O \O' \a ) ] = \E[ f( \tilde\a' (\O' \z), \O \tilde\a )]  =
 \E[ f(\a' (\O' \z), \O \a ] $ where in the last equality we replaced $\tilde\a$ by $\a$ for simplicity. It is valid since both have the same distribution.  If we pick $\O$ as $\O = [\frac{\z}{\|\z\| }, \O_{rest}]$ with $\O_{rest}$ being anything that ensures $\O$ is unitary and independent of $\a$, this simplifies to $ \E[ f( a_1 \|\z\|, \O \a ]$. Here $a_1$ is the first entry of $\a$.   This expectation is much easier to compute.
 As an example, when trying to be compute $\E[\Y_0]$ for $\Y_0$ defined in \eqref{def_Y0} above, $\z \equiv \xstar$ and $f(x_1,\x_2) \equiv x_1^2 \x_2 \x_2{}'$. Using above,  $\E[\Y_0] = \|\x\|^2 \E[a_1^2 \O \a \a' \O'] = \|\x\|^2 \O\E[a_1^2  \a \a' ] \O'$. It is  easy to see that $\E[a_1^2  \a \a' ] = c_1 \I + c_2 \e_1 \e_1{}'$ and thus $\|\x\|^2 \O ( c_1 \I + c_2 \e_1 \e_1{}' ) \O' = c_1 \|\x\|^2 \I + c_2 \xstar \xstar{}'$ for scalars $c_1, c_2$. Here $\e_1$ is the first column of the identity matrix $\I$.

The above idea can be extended if instead of one vector $\z$, there are a few vectors, all independent of $\a$. To understand the idea simply, suppose there are three such vectors $\z_1, \z_2, \z_3$ and we need to compute $\E[ f ( \a'\z_1, \a'\z_2, \a'\z_3, \a  )]$. Then one can pick $\O$ so that the span of its first three columns equals that of $[\z_1, \z_2, \z_3]$. With doing this,  $\E[ f ( \a'\z_1, \a'\z_2, \a'\z_3, \a  )] = \E[ g ( a_1, a_2, a_3, \O \a )]$ which is much easier to compute or bound than the original expression, especially if $f( )$ is such that $\O$ can be pulled out of the expectation sign as in the above example.
Assuming ``sample splitting'' described above allows one to use the above approach to deal with functions of $\a' \xhat$, where $\xhat$ is the estimate from the previous iteration.

Finally, if a structured PR problem involves recovery from non-global measurements, the distribution of the different measurements conditioned on the signal can be very different. For example, this is the case for low rank PR. However in order to apply the concentration bounds on functions of all the measurements, one needs the distributions to be ``similar enough''. For recovering low-rank matrices, one solution is to assume ``incoherence" of its left or right singular vectors just as is done for solving LRMC. 

\cbl


\section{Sparse and Low Rank PR}

{\em We provide a summary of the guarantees for the various solution approaches for sparse and low-rank PR and for linear problems that are most related to these problems in Table \ref{compare_assu}.}

\subsection{Sparse PR Problem}
Sparse PR involves recovering an $s$-sparse signal $\xstar$ from $m$ phaseless linear projections $\y$ satisfying \eqref{y_mod}.
Let $\T$ denote the support set of $\xstar$. By assumption $|\T|\le s$. Sparse PR can be simply understood as a phaseless (magnitude-only) version of Compressive Sensing (CS).  Like CS, it also involves recovery from global measurements -- each measurement $\y_i$ is a function of the entire unknown vector $\xstar$. 

\subsection{Low Rank PR: Two Problem Settings}
The low-rank assumption can be used in one of the following two ways.

\subsubsection{A single signal can be reshaped into a low-rank matrix}
This assumes that the unknown signal or image, whose phaseless linear projections are available, can be reshaped to form a low-rank matrix.
To be precise, one assumes that the unknown $nq$-length signal $\xstar$ is such that it can be reshaped to form an $n \times q$ matrix $\Xstar$ that has low rank $r$. The goal is to recover $\Xstar$ from measurements $\y_i:= | \langle \A_i, \Xstar \rangle |, i=1,2,\dots, m$. This problem can be understood as the phaseless version of LRMS.
%
%
%
This model is valid only for very specific types of images for which different image rows or columns look similar, so that the entire image matrix can be modeled as low rank. An example is images of textures, e.g. green spaces with no foregrounds. In general it is not a very practical model, and this is probably why {\em this setting has not been explored in the literature.}

\subsubsection{A set of signals form a low-rank matrix (Low Rank PR)}
A more practical model, and one that is commonly used in many biomedical applications \cite{st_imaging}, is to consider the dynamic imaging setting and assume that a set, e.g., a time sequence, of signals/images is generated from a lower dimensional subspace of the ambient space \cite{lrpr_tsp,lrpr_icml,lrpr_it}. We have $m$ different and independent phaseless linear projections of each signal. The question is when can we jointly recover the signals using an $m \ll n$?
Said another way, the goal is to recover an $n \times q$ matrix $\Xstar$ which has rank $r$  from 
\begin{align}\label{obsmod}
\y_{ik} := | \langle \a_{ik} \bm{e}_k{}', \Xstar \rangle| =  | \langle \a_{ik}{}, \xstar_k \rangle|, \ i = 1,\ldots, m, \ k = 1,\ldots,q.
\end{align}
when all the $\a_\ik$'s are mutually independent. This model can also be rewritten as $\y_k:= |\A_k \xstar_k|, k=1,2,\dots,q$ with $\A_k:=[\a_{1k}, \a_{2k}, \dots, \a_{mk}]'$.
 Thus the total number of available measurements in this case is $mq$.
Here $\xstar_k$ is the $k$-th signal ($k$-th column of $\Xstar$) and $\e_k$ is the $k$-th column of the identity matrix, $\I$.  This problem formulation, dubbed Low Rank PR (LRPR), is valid any time the set/sequence of signals is sufficiently correlated so that the differences between the $q$ different images can be explained by only $r$ linearly independent factors. It is useful, for example, for low-cost fast phaseless dynamic imaging, e.g., dynamic Fourier ptychography (ptychographic imaging of gradually changing dynamic scenes, such as live biological specimens, in vitro). In order to capture changes in the scene, the image acquisition process needs to be fast enough; this can be enabled if one can obtain accurate recovery using fewer measurements/samples.

Observe that \eqref{obsmod} uses a different set of $m$ measurement vectors, $\a_\ik, i=1,2,\dots,m$ for each signal/column $\xstar_k$. As we explain with a simple example, this is {\em necessary} to allow correct recovery using an $m < n$. Consider the $r=1$ setting and suppose that $\xstar_k = \xstar_1$ (all columns are equal). With the above set up, we then have $mq$ i.i.d. Gaussian measurements of $\xstar_1$ and hence, by an standard PR result, $mq \in \Omega (n)$ suffices. If $q \approx n$, this means that $m \in \Omega(1)$ suffices. 
On the other hand, if $\a_\ik = \a_i$ for all $k$, then in this case, only the first $m$ measurements are useful, the others are just repeats of these. Thus, we will end up needing $m \in \Omega (n)$ making the sample complexity as high as that of standard PR. This $\a_\ik = \a_i$ case, and its linear version, is what has been studied extensively in the literature \cite{wainwright_linear_columnwise,cov_sketch}. For this case, $m = nr$ is in fact necessary.


Notice that in  \eqref{obsmod}, we have global measurements of each column of $\Xstar$, but not of the entire matrix. Thus, in order to correctly recover $\Xstar$ (using  an $m < n$), we need an assumption that allows for correct ``interpolation'' across the rows. One way to ensure this is to borrow the ``incoherence of right singular vectors'' {\em (right incoherence)} assumption from the LRMC literature \cite{nonconvex_review}. 
Modulo constants (and assuming that the condition number is a numerical constant), this assumption can be understood simply as  requiring that \cite{lrpr_it} 
\begin{align}
\max_k \|\xstar_k\|_2^2   \leq  C  \frac{\|\X^*\|_F^2 }{ q}.
\label{right_incoh_2}
\end{align}



\input{table}


\subsection{Sparse PR Solutions and Guarantees}
\cblue
Sparse PR approaches can be split into four categories: (i) convex relaxation approaches such as $\ell_1$-PhaseLift  \cite{voroninski13} or regularized PhaseMax; (ii) older methods for Fourier sparse PR \cite{jaganathan2013sparse2} that use a combinatorial algorithm for support estimation followed by convex relaxation; (iii) a series of provably correct and fast iterative approaches for solving standard sparse PR ($\a_i$'s are i.i.d. random Gaussian): AltMinSparse \cite{pr_altmin}, Sparse Truncated Amplitude Flow (SPARTA)  \cite{sparta}, Thresholded WF \cite{cai} and CoPRAM \cite{fastphase}; and (iv) approaches that assume that one can choose a different design for the measurement vectors \cite{ramachandran15,bahmani}. $\ell_1$-PhaseLift is similar to PhaseLift and again involves lifting the problem to an $n^2$ dimensional space to make it convex: one attempts to recover the matrix $(\x \x')$ by imposing the data constraint and requiring the matrix to be both sparse and low rank (ideally it should be rank one). It is thus very expensive both in time and space complexity. It needs sample complexity  $m \in \Omega (s^2 \log n)$ and was the first provably sample-efficient solution to sparse PR. Regularized PhaseMax is the PhaseMax idea applied for a structured PR such as sparse PR. So it is a convex relaxation solution that is about as computationally efficient as the non-convex approaches. However the guarantee for it is asymptotic.
Since this review focuses on non-convex solutions to standard structured PR, we only describe approaches in category (iii) in detail.

\cbl

\subsubsection{Non-convex standard sparse PR approaches}
AltMinSparse was the first such method \cite{pr_altmin}. This was developed as an extension of the first provable non-convex solution to standard PR, AltMinPhase. In the initialization step, it obtains a one-shot estimate, $\That$, of the support of the sparse vector
 as the set of indices of  the $s$ largest magnitude diagonal entries of the matrix $\Y_0$ defined in \eqref{def_Y0}. Since $\E[\Y_0] = \|\xstar\|_2^2 \I + 2 \xstar \xstar{}'$, thus $(\E[\Y_0])_{j,j} = \|\xstar\|_2^2 + 2 (\xstar)_j^2$. Thus, by assuming that $\Y_0$ is close to its expected value (true w.h.p. if $m$ is large enough), and the smallest nonzero entry of $\xstar$, denoted $\xstar_{\min}$, is large compared to the bound on $\| \Y_0 - \E[\Y_0]\|_2$, one can show that $\That$ will be equal to the true signal support $\T$.
After this, AltMinSparse just implements the AltMinPhase algorithm, but for recovering the $|\That|$-length signal $\x_{\That}$. The signal estimate is set to zero on the complement set, $\That^c$. 

In follow-up parallel work, two other algorithms were developed and studied: Sparse TAF (SPARTA)  \cite{sparta} and Thresholded WF (ThreshWF) \cite{cai}. SPARTA uses the same exact approach as AltMinSparse with the only difference being that AltMinPhase is replaced by TAF \cite{taf}. Thus it is a GD approach.
Since both AltMinPhase and SPARTA estimate the support only once in the beginning, both require a lower bound on the smallest nonzero entry of $\xstar$ (denoted $\xstar_{\min}$).

ThreshWF \cite{cai} is more interesting because, unlike AltMinSparse or SPARTA, it updates the signal support estimate at the end of each iteration. This simple but important modification helps remove the (unnecessary) lower bound on $\xstar_{\min}$ that AltMinPhase and SPARTA need. Briefly, ThreshWF involves a spectral initialization followed by a projected GD type algorithm: each iteration involves one step of GD to minimize \eqref{costfn} followed by a thresholding step that makes the new estimate sparse. The threshold itself is carefully computed at each iteration using the previous signal estimate. 
\cblue
Spectral initialization consists of: (i) compute the matrix $\Y_0$ defined above, (ii) estimate the signal support, $\That$, as the set of indices of its $s$ largest magnitude diagonal entries, and (iii) compute $\xhat_{\That}$ as the top eigenvector of the sub-matrix $(\Y_0)_{\That,\That}$.
\cbl

%
In \cite{fastphase}, an AltMin approach for sparse PR is developed. After spectral initialization similar to the one above for ThreshWF, it alternates between estimating the measurements' phase, and solving a compressive sensing (CS) problem to estimate the support and signal values of the sparse vector using the measurements multiplied by the estimated phase as a proxy for the linear measurements in CS.
Like ThreshWF, since this approach also updates the support estimate at each iteration (done while solving the CS problem), it also does not require a lower bound on $\xstar_{\min}$.  
Time complexity of both the approaches is roughly similar too:  $C m n \log(1/\epsilon)$ and  $C m n \log^2(1/\epsilon)$ respectively.
%


\subsubsection{Non-convex standard sparse PR guarantees}
The first two fast non-convex approaches for standard sparse PR -- AltMinSparse and SPARTA -- needed to assume a lower bound on the minimum nonzero entry of $\x$. In follow-up work on Thresholded WF and then later on CoPRAM, this extra assumption was removed because these last two approaches updated the support estimate of the signal at each iteration. All four results need $m = \Omega (s^2 \log n)$ measurements. The time complexity of SPARTA and ThreshWF was $\bigo(mn \log(1/\epsilon) ) =  \bigo(  n s^2 \log n \log(1/\epsilon)) $ if we replace $m$ by its lower bound. On the other hand, since AltMinSparse and CoPRAM are AltMin approaches, the LS or CS steps in each main iteration involved using an iterative algorithm (e.g. Conjugate Gradient LS for LS or CoSaMP for CS). Because of this, their time complexity has one more $\log(1/\epsilon)$ factor. Thus their time complexity is $\bigo(mn \log^2(1/\epsilon) )$.


\subsection{Low Rank PR Solutions and Guarantees}
The LRPR problem was introduced and first studied in \cite{lrpr_tsp} where we developed two simple algorithms -- LRPR1 and LRPR2, evaluated them experimentally, and provided a guarantee for their initialization step. In later work \cite{lrpr_icml,lrpr_it},we proposed a significantly improved algorithm,  Alternating Minimization for Low Rank Phase retrieval (AltMinLowRaP), for which a complete correctness guarantee could be proved. In a recent preprint \cite{lrpr_improved} we have improved this guarantee by a factor of $r$ and have also obtained a result for complex Gaussian measurements.

\subsubsection{Solutions}
AltMinLowRaP \cite{lrpr_icml,lrpr_it} uses the factorization of a rank $r$ matrix $\X$ into $\X = \checkU \checktB$ where $\checkU$ is $n \times r$ and $\checktB$ is $r \times q$. It minimizes
\begin{align} \label{optprob}
\sum_{k=1}^q \| \  \y_k -  |\A_k{}' \checkU \checktb_k| \ \|_2^2  
\end{align}
alternatively over $\checkU, \checktB$ with the constraint that $\checkU$ is a tall matrix with orthonormal columns.
At a top level, the AltMin can be understood as alternating PR: minimize \eqref{optprob} over $\checktB$ keeping $\checkU$ fixed at its current value and then vice versa. But there are important differences between the two PR problems and how they can be solved.
Given the previous estimate of $\Span(\Ustar)$, denoted $\hat\U$, the recovery of each $\tb_k$ is an easy $r$-dimensional standard unstructured PR problem. This is true assuming ``sample-splitting'' (a different set of $mq$ measurements is used for each update of $\Ustar$ and another new one for each update of $\tb_k$'s). 
%
Given an estimate of $\tB$, denoted $\hat\B$, the update of $\Ustar$ (or equivalently of its vectorized version, $\Ustar_{vec}$) is a  significantly non-standard PR problem for two reasons. First, the measurement vectors are no longer independent or identically distributed. Second, by using the previous estimates of $\Ustar$ and of $\tb_k$, with accuracy level $\delta$, we can get an estimate, $\xhat_k = \hat\U \bhat_k$, of $\xstar_k$ with the same accuracy level. With this, we can also get  an estimate of the phase of the measurements, $\cb^*_\ik:= \mathrm{phase}(\a_\ik{}' \xstar_k)$ with the same accuracy level. As a result, obtaining a new estimate of $\Ustar_{vec}$ becomes a much simpler Least Squares (LS) problem rather than a PR problem\footnote{As explained in \cite{lrpr_it}, a similar argument does not apply when recovering $\tb_k$'s. The reason is, with a new estimate of $\Ustar$, denoted $\U^+$, the previous estimate of $\tb_k$ becomes useless: it is close to $\U'\Ustar \tb_k$ in phase-invariant distance, but not to $\U^+{}' \Ustar \tb_k$.}. 

A simpler way to understand the approach is to split it into a three-way AltMin problem over $\Ustar$, $\tb_k$, and $\cb^*_\ik:=\mathrm{phase}(\a_\ik{}' \xstar_k)$. Given an estimate of $\Ustar$, recover $\tb_k$'s by solving $r$-dimensional standard PR problems; estimate $\hat\cb_\ik= \mathrm{phase}(\a_\ik{}' \Uhat \bhat_k)$; and then update $\Uhat$ by solving the following LS problem: $\arg\min_{\tilde\U} \sum_{k =1}^q\| \Chat_k \y_k^{(T+t)} - \A_k^{(T+t)}{}' \tilde\U \b_{k}^{t}\|_2^2$ followed by QR decomposition on the solution.

The initialization step adapts the truncated spectral initialization idea of \cite{twf} for solving LRPR. One obtains an initial estimate of $\Span(\Ustar)$ by computing the top $r$ eigenvectors of
\beq \label{def_YU}
\Y_U := \frac{1}{mq} \sum_{k=1}^q \sum_{i=1}^m \y_{ik}^2 \a_{ik}\a_{ik}' \indic_{ \left\{ \y_{ik}^2 \leq 9 \kappa^2 \mu^2 \frac{1}{mq}\sum_{ik} \y_{ik}^2 \right\}  }.
\eeq
One can show that $\E[\Y_U] = t_1 \I + t_2 \Ustar (\tB \tB{}) \Ustar{}'$ for scalars $t_1, t_2$, and thus its top $r$ eigenvectors span $\Span(\Ustar)$. Truncation (throwing away $\y_\ik$'s that are too large compared to their empirical mean) helps ignore  the``bad'' measurements that can incorrectly bias $\Y_U$ and hence its singular vectors. Here $\indic_{\text{statement}}$ denotes the indicator function: it is equal to one if the statement is true and zero otherwise.

LRPR1 from  \cite{lrpr_tsp} was a projected GD, or to be precise, a projected truncated GD solution. Its initialization step estimated $\Ustar$ using the above approach, and then estimated $\bstar_k$'s using a standard PR spectral initialization. At each iteration, it involved one step of GD for each column $\xstar_k, k=1,2,\dots,q$ (implemented using one iteration of truncated WF \cite{twf}), followed by projecting the resulting matrix onto the space of rank $r$ matrices.

\subsubsection{Guarantee}
Our guarantee \cite{lrpr_icml,lrpr_it} shows that AltMinLowRaP will recover each column of $\Xstar$ to $\epsilon$-accuracy in normalized phase-invariant distance, $\sum_{k=1}^2 \dist^2(\xhat_k, \xstar_k) / \|\Xstar\|_F^2$,  as long as
\bi
\item the right incoherence assumption given in \eqref{right_incoh_2} holds;
\item we use sample-splitting; and
\item the total measurements per column, $m$, satisfies $m \in \Omega( \frac{n}{q} r^4 \log (1/\epsilon) )$ and    $m \in \Omega ( \max(r,\log n, \log q) ) \log (1/\epsilon) )$.
\ei
The second requirement on just $m$ is redundant except when $q$ is very large, $q > nr^3$.

In more recent work \cite{lrpr_improved}, we have been able to improve the above guarantee by a factor of $r$ to only require
\bi
\item $m \in \Omega( \frac{n}{q} r^3 \log (1/\epsilon) )$ in addition the second (usually redundant) requirement.
\ei
When $q \ge n$, this guarantee means that only about $r^3$ measurements per signal suffice. When $r$ is small, this is a significant reduction over unstructured PR which necessarily needs $m \ge n$. This reduction is possible because, for both the initialization and the update steps for updating $\Ustar$, we have access to $mq$ mutually independent measurements. These are not identically distributed (because the different $\bstar_k$'s could have different distributions), however, one can carefully use the right incoherence assumption to show that the distributions are ``similar enough'' so that the concentration inequality (sub-exponential Bernstein) can be applied jointly for the sum over all $mq$ samples or their functions.


\subsubsection{Other possible solution approaches}
When solving a structured PR problem, one can design the AltMin approach in more than one way. There are 3 sets of variables:  the phase of each measurement, the structure (the column span matrix $\Ustar$ in case of a low-rank matrix and the support $\T$ in case of a sparse vector),  and the coefficients ($\Bstar$ in case of low-rank matrix and nonzero entries of $\xstar$ in case of sparse vectors).
The AltMinLowRaP solution for Low Rank PR alternates between (i) updating the coefficients $\Bstar$ and  the measurements' phases and (ii) updating the structure $\Ustar$.
However, a different approach can also be developed  that is similar in principle to the AltMin approach for sparse PR: one can alternate between (i) estimating the measurements' phases, and (ii) solving the linear version of low-rank PR (compressive PCA). This approach can be analyzed easily by borrowing the phase error term bound from the AltMinLowRaP guarantee from \cite{lrpr_improved}. When analyzed this way, the guarantee for this new approach is exactly the same as that for AltMinLowRaP.

LRPR1 from \cite{lrpr_tsp} is the projected GD solution. It has so far not been analyzed theoretically. The AltminLowRaP guarantee cannot be improved beyond the current value, thus, if would like to reduce complexity to $(n/q)r^2$ we should try to analyze LRPR1.

Lastly it is possible to extend the PhaseMax idea to low rank PR as well. The design is easy but the analysis remains an open question.

\subsection{Sparse versus Low Rank PR}
Sparse PR involve recovery from global measurements of the sparse signal $\xstar$.  LRPR involves recovery from measurements that only depend on individual columns of $\Xstar$ and not on the entire $\Xstar$. As is well known both from the CS and the low-rank matrix recovery literature, the global measurements' setting is easier to solve (one can obtain better sample complexity guarantees) than its non-global counterpart since one can often prove a restricted isometry property (RIP) for it. For example, the best  sample complexity guarantee for LRMS (which has global measurements) is near optimal while that for LRMC (local measurements) is $r$ times sub-optimal.

Non-global measurements is what makes LRPR a more difficult problem than sparse PR.
The best existing Low Rank PR guarantee needs $m \in \Omega ((n/q) r^3 \log (1/\epsilon) )$, while the best one for sparse PR (phaseless but global measurements) needs $m \in \Omega (s^2 \log n)$.

\subsection{Most closely related problems that are well studied}
The linear (with phase) version of sparse PR is Compressive Sensing (CS).  The best sample complexity for CS with Gaussian measurements requires $m \in \Omega (s \log n)$ while that for sparse PR, which is a harder problem, is $m \in \Omega (s^2 \log n)$.

%
While one would think that the linear (with phase) version of Low Rank PR, recover $\Xstar$ from $\y_\ik:= \langle \a_\ik \e_k{}', \Xstar \rangle, i=1,2,\dots, m, k=1,2,\dots q$, would been extensively studied, this is not true. This problem is typically called ``Compressive PCA''  or ``PCA via random projections''.
\cblue
There have been some older attempts to develop a solution and try to analyze sub-parts of it \cite{hughes_icml_2014, aarti_singh_subs_learn}. In fact, AltMinLowRaP can be understood as the first provably correct algorithm for this problem as well. In very recent work \cite{lee2019neurips}, a provable convex optimization (mixed norm min) approach was developed. Its guarantee needs right incoherence and  $mq > \frac{r (n+q) \log^6 (n+q) }{ \epsilon^2}$ to achieve $\epsilon$ accuracy. Speed-wise this is significantly slower than AltMinLowRaP which is a non-convex solution. Its sample complexity is better or worse depending on the choice of $\epsilon$. When $\epsilon < 1/r$, ignoring log factors, the AltMinLowRaP sample complexity is better, otherwise that of mixed norm min is better.

The closest linear low rank recovery problem to LRPR that is well-studied is LRMC. LRMC involves recovery from row-wise and column-wise local measurements while LRPR measurements are row-wise local but column-wise global. Ignoring log factors, the best existing Low Rank PR guarantee needs $mq$ of order $n r^3$, while that for LRMC needs $mq $ of order $n r^2$ \cite{rmc_gd}. An open question is whether the projected GD solution to LRPR (LRPR1 \cite{lrpr_tsp}) can be analyzed to achieve $nr^2$ sample complexity?
\cbl

\section{Which Model is Better When and Experimental Evaluations}

A common question when using structural assumptions for real datasets is which one is the best one to use and when?

\subsection{Sparsity or Low-Rank}
Sparsity (and certain types of structured sparsity) are simpler assumptions that are applicable in both the static (single image) and the dynamic (image sequence) imaging settings. Moreover, fast algorithms with low sample complexity guarantees are easier to develop for sparse PR.
On the other hand, as explained earlier, modeling a single image as being low-rank is typically not a practical assumption.  Of course, even if it were practical, there is no existing solution so far for exploiting it.

Low-rank is often a much more reasonable modeling assumption for dynamic imaging applications involving a time sequence of similar (correlated) signals/images, e.g., bacteria growing a petri dish. This is the LRPR setting described above.
In this case, the low rank prior is a significantly more flexible one than sparsity or structured sparsity. The reason is that it does not require knowledge of the dictionary or basis in which the signal is sufficiently sparse. For example, this is demonstrated in Fig. \ref{bact_comp}  below.
As can be seen, when using a generic sparsity basis like wavelet (it is well known that all piecewise smooth images will be sparse to some extent in the wavelet basis),  the recovery performance is much worse than just using the low rank assumption.

It should be mentioned that low-rank includes certain types of dynamic sparsity models (those with fixed or small number of changes in support over time) as special cases \footnote{For readers familiar with low rank matrix completion, this point may cause some confusion. LRMC requires denseness of both left and right singular vectors. Denseness of left singular vectors implies that the columns of the unknown matrix cannot be sparse. As a result, one cannot complete a matrix that is both low-rank and sparse using a subset of its entries. This is not true in case of LRPR because LRPR involves column-wise dense measurements and thus does not require denseness of left singular vectors.}.

The drawback of using low-rank for dynamic imaging is that, for a given value of $m$, it often results in slower approaches than those for unstructured PR. 



%
\subsection{Experimental comparisons for Fourier Ptychography} \label{expts}
Due to limited space, we only show one set of experimental comparisons for one real application: Fourier ptychography. Fourier ptychography is an imaging technique used in both microscopy and long-distance imaging to mitigate the effects of diffraction blurring. It uses an array of images from low-resolution cameras to produce a high-resolution image. In signal processing language, each camera image corresponds to the magnitude of a different band-pass filtered version of the target high-resolution image.  To get enough measurements per image, one either needs many cameras (expensive), or one needs to move a single camera to different locations to acquire the different bands~\cite{holloway}. This can make the acquisition process slow or expensive and hence approaches that help reduce sample complexity can be useful.

In order to be able to report quantitative results on realistic data, we use real images or videos (these are only approximately sparse or low-rank),  but we simulate the ptychographic measurement acquisition set up, this uses \cite[equation (3)]{TCIgauri}.  %

{\em Static imaging: Comparing Sparse PR and unstructured PR solutions. }
Static imaging cannot be posed as an LRPR problem. Thus we only compare the various sparse PR solutions. We use the resolution chart image shown in Fig. \ref{reso_comp}(a) for the comparisons. It contains enough fine details which make it an interesting image to evaluate structured PR solutions. We compare SPARTA, CoPRAM, and IERA (Iterative Error Reduction algorithm) which is an (unstructured) PR solution for Fourier ptychography \cite{holloway}. For solving the Fourier ptychographic imaging problem, the algorithms need to be initialized differently; one needs to use the approach eexplained in Algorithms 1 and 2 of \cite{TCIgauri}.
 We display the {\em structural similarity index (SSIM)} between the recovered and true image in Fig. \ref{reso_comp}(b) for four undersampling levels. SSIM is a measure of normalized cross-correlation between two images and thus higher SSIM implies better reconstruction. This experiment is borrowed from \cite{TCIgauri}.

{\em Dynamic Imaging: Comparing Low-Rank PR, Sparse PR and unstructured PR solutions. }
For dynamic imaging, as explained earlier both sparse and low-rank PR are applicable. 
%
We used a few real slow changing videos and simulated dynamic Fourier ptychography measurements. We show comparisons on one such video (the bacteria video shown in Fig. \ref{bact_comp}(a)) here. We compared AltMinLowRaP (an LRPR solution), CoPRAM (a sparse PR solution), block-CoPRAM (a structured sparse PR solution), and IERA (an unstructured PR solution for Fourier ptychography) \cite{holloway}.
Since the video is not exactly low-rank, we implemented AltMinLowRaP with a modeling error correction step. This step applies a few iterations of any standard PR approach, here IERA, column-wise to the output of AltMinLowRaP, in order to also estimate some of the ``modeling error'' in the low-rank assumption. We display the structural similarity index (SSIM) comparison between the estimated and true videos in Fig. \ref{bact_comp}(b) for four undersampling levels. SSIM is a measure of normalized cross-correlation between two images and thus higher SSIM means better reconstruction. This experiment is borrowed from \cite{TCIgauri} and \cite{ICIP20}.

\section{Other Structural Assumptions} 

\subsubsection{Bayesian priors}
Structural assumptions or the above dynamic extensions can also be imposed via Bayesian priors. But Bayesian priors need a large number of model parameters and there is often a model-mismatch between the model learned using training data and the model followed by the test data. Of course, when it is valid to assume that test and training data are generated from the same distribution, Bayesian priors enable better reconstructions than non-Bayesian ones.

\subsubsection{Deep Neural Network prior}
A recent work assumed a deep neural network based ``generative prior'' on the signal $\xstar$  \cite{pr_generative_prior}. To be precise, it assumed that the $n$-length signal/image $\xstar$ lies in the range space of a trained $d$-layer feed-forward neural network, with Rectified Linear Unit (ReLU) activation, whose input is an unknown $s$ length vector with $s \ll n$. If we use $G$ to denote such a neural network, then we are assuming that the signal/image to be recovered, $\xstar$, can be modeled as $\xstar := G(\b)$ where $\b$ is an $s$-length vector and $s \ll n$.  The goal is to recover $\b$, and hence $\xstar$, from $\y:= |\A \xstar| = |\A G(\b)|$. Here $\A$ is the $m \times n$ measurements' matrix. The recovery algorithm is GD. Under this model, the authors show that if $m = \Omega (s d^2 \log n)$, and if certain other assumptions on $G$ hold, then, the problem setting has ``favorable global geometry for gradient methods''. This work follows up on previous work that solves the compressive sensing problem (linearized sparse PR) using a similar deep neural net based generative prior. 

\subsubsection{Compression prior}
In another recent work \cite{maleki_isit18} a different generative assumption has been used on $\x$: it assumes that $\x$ can be compressed with rate-distortion function $r(\delta)$ and uses this assumption to develop an efficient compressive PR algorithm. Thus the prior assumed is that the signal can be compressed with small distortion using a given compression scheme.



\subsubsection{Structure dynamics over time}
One can further improve sparse or low-rank PR for dynamic imaging by exploiting structure dynamics over time. These have been extensively explored for three other related problems -- CS, LRMC, and Robust PCA. Dynamic sparsity assumptions, such as slow signal support change over time, have been shown to significantly improve the performance of dynamic CS algorithms \cite{dyncs_review}. Slow support change is often a valid assumption for slow changing sparse signal or image sequences that often occur in biological or medical imaging applications. Similarly, dynamic low rank assumptions such as slow subspace change have significantly improved the state of the art for dynamic robust PCA  (robust subspace tracking) and dynamic LRMC. 
Dynamic low-rank has also been briefly explored in the PR literature in \cite{lrpr_icml,lrpr_it}.
%
Dynamic low-rank is a good idea only for long time sequences of datasets while dynamic sparsity can be useful even for shorter sequences. 

\input{figures_2}

\section{Open Avenues for Future Work}
There are a large number of open questions for future work in this area of structured quadratic sensing or phase retrieval. A difficult, and as yet unsolved open question (even after many years of work on the topic) is whether the sample complexity of sparse PR can be reduced from $s^2 \log n $ to $s \log n$.
Moreover the low rank PR problem has not received as much attention. Open sample complexity questions include (i) can the sample complexity be reduced by another factor of $r$ to $(n/q)r^2$ times log factors (comparable to the best existing non-convex LRMC guarantee), and (ii) can we  remove the need for using a new independent set of measurements at each iteration? It may be possible to tackle (ii) using leave-one-out ideas similar to \cite{pr_mc_reuse_meas}. In order to address (i), the projected GD approach (LRPR1 from \cite{lrpr_tsp}) should be studied. As explained in \cite{lrpr_improved}, one cannot improve the sample complexity of AltMinLowRaP any further. 
(iii) Another related question of interest is whether one can obtain the above sample complexity guarantee by designing a  PhaseMax-based approach for LRPR.



A question of both practical and theoretical interest is how to  come up with a simple and fast algorithm for solving the problem of ``sparse and low-rank'' PR and whether such a method improves upon using either just sparsity or just low-rank in practical phaseless dynamic imaging settings. This type of modeling been used very successfully in the MRI literature to get the best possible sample complexity in empirical experiments \cite{dyn_mri1}. It has also been studied theoretically in recent work for the linear low-rank and sparse matrix recovery problem \cite{lee2017near}.
A third  practically important question is how to develop PR approaches that exploit dynamic sparsity (e.g., slow support change) or dynamic low-rank (e.g., slow subspace change). 
%
Finally, the phaseless LRMS problem is another problem that is as yet unsolved.

On the applications end, structured PR methods can be exploited in many newer domains such as crystallography, astronomy, and optics, where these methods have not been explored carefully so far.
\cblue
A relevant theoretical question is how to analyze structured PR approaches that use masked Fourier or other Fourier-based measurement models that are actually used in practical applications of PR.
\cbl

\bibliographystyle{IEEEbib}
\bibliography{../../../Proposals/PR-CCF-Small-Nov15/CH/chinbiblio,../../../Proposals/PR-CCF-Small-Nov15/CH/biblio_nips,../../../bib/tipnewpfmt_kfcsfullpap}


\end{document}

%% file: table.tex
\begin{table*}[t!]
\caption{\small{Comparing non-convex solutions for sparse and low-rank PR. Here ``best'' means best sample complexity. As baseline, we also list best results for the most related problems that are well studied. The comparison treats $\kappa,\mu$ as constants. $\Xstar$ is an $n\times q$ rank-$r$ matrix; $\xstar$ is an $n$-length $s$-sparse vector. CS and LRMS are linear problems with global measurements and thus their sample complexity guarantees for non-convex solutions are near-optimal. Sparse PR has global but phaseless measurements, LRMC has linear but non-global measurements and LRPR has both phaseless and non-global measurements. 
}}
\begin{center}
\renewcommand*{\arraystretch}{1.01}
\resizebox{\linewidth}{!}{
\begin{tabular}{|l|l|l|l|l|} \toprule \hline
Problem & Global & Assumptions & Sample Complexity  & Time Complexity per signal  \\
& Measurements?  &&  $m \ge C \cdot $  & (with $m$=its lower bound)  \\ \hline 
%
%
Sparse PR & Yes  & $\xstar$ is $s$-sparse, & $ s^2 \ \log n \log (\frac{1}{\epsilon})  $ &  $\bigo(  m n s  \log (\frac{1}{\epsilon}) ) = $ \\
 (first) \cite{pr_altmin} &     &              $\xstar_{\min}$ lower bounded & & $ \bigo( n s^3 \log n \log^2(\frac{1}{\epsilon}) )$ \\ \hline
Sparse PR &  Yes  & $\xstar$ is $s$-sparse & $ s^2 \  \log n $ &  $ \bigo( m n  \log(\frac{1}{\epsilon}) ) = $ \\
 (best) \cite{cai} &&&&  $\bigo(  n s^2 \log n \log (\frac{1}{\epsilon}) )$  \\ \hline
Comp. Sens. &  Yes  & $\xstar$ is $s$-sparse & $  s \  \log n  $ &  $\bigo(  m n  \log(\frac{1}{\epsilon}) )= $ \\
(best)  &&&&  $\bigo( n s  \log n \log(\frac{1}{\epsilon}) )$  \\ \hline
LRPR&  No & $\Xstar$ has rank $r$  & $ \frac{n}{q} r^4 \ \log (\frac{1}{\epsilon}) $   &    $\bigo(  m n r  \log^2 (\frac{1}{\epsilon}) )= $   \\
 (first)   \cite{lrpr_icml}  &  &   right incoherence   && $\frac{1}{q} \bigo( n^2  r^5  \log^2(\frac{1}{\epsilon}) )$ \\ \hline
LRPR&  No & $\Xstar$ has rank $r$  & $ \frac{n}{q} r^3 \  \log (\frac{1}{\epsilon}) $   &    $\bigo(   m  n r  \log^2 (\frac{1}{\epsilon}) )= $   \\
 (best so far)   \cite{lrpr_improved}  &  &   right incoherence   && $\frac{1}{q}\bigo(  n^2  r^4  \log^2(\frac{1}{\epsilon}) )$ \\ \hline
LRMC  & No  & $\Xstar$ has rank $r$, & $\frac{n}{q} r^2  \ ( \log n)^2 (\log (\frac{1}{\epsilon}) )^2 $ &  $\bigo( m  r   \log^2 ( \frac{1}{\epsilon}) )= $ \\
 (best)  &   & left \& right incoherence  &&  $\frac{1}{q} \bigo( n r^3 \log n \log^3( \frac{1}{\epsilon}) )$ \\ \hline
LRMS & Yes  & $\Xstar$ has rank $r$  & $\frac{n}{q} r \  \log n  $ &  $\bigo(  m  n r  \log^2 (\frac{1}{\epsilon}) ) = $ \\
 (best)   &   &   &&  $\frac{1}{q}  \bigo(  n^2 r^2 \log n  \log^2 (\frac{1}{\epsilon}) )$  \\ \hline
\bottomrule
\end{tabular}
}
\label{compare_assu}
\end{center}
\end{table*}

%% file: figures_2.tex
\begin{figure}[t!]
\subfigure[Original image]{
\centering
			{\includegraphics[width = 0.3\textwidth]{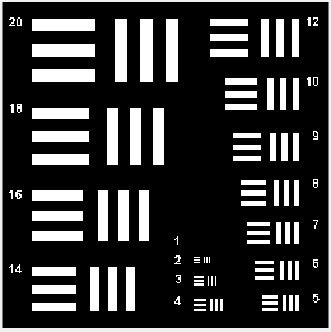}}
}
\subfigure[Recovery SSIM plotted against undersampling level]{
\centering
	\input{plot.tex}
}
	\caption{\small{{\bf Comparing various sparse PR solutions and IERA (unstructured PR solution for Fourier ptychography) \cite{holloway} on the resolution chart image shown in (a):}
 This figure is taken (re-plotted using data) from \cite{TCIgauri}.
}}
\label{reso_comp}
\end{figure}

		\begin{figure}[t!]
\subfigure[Three frames of original bacteria video]{
\centering
\begin{tabular}{ccc}
		{\includegraphics[width = 0.2\textwidth]{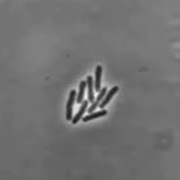}} &
		{\includegraphics[width = 0.2\textwidth]{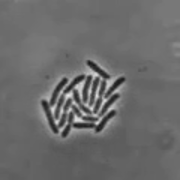}} &
		{\includegraphics[width = 0.2\textwidth]{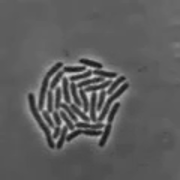}} \\
		frame 30 & frame 66 & frame 90 \\
\end{tabular}
}
\\
\subfigure[Recovered video SSIM plotted against undersampling level]{
\centering
		\begin{tikzpicture}[scale=1.25] 
\centering
		\begin{axis}[xlabel=$f$,ylabel=SSIM,grid style = dashed, ylabel near ticks, xlabel near ticks, grid=both,legend columns=2,legend style={/tikz/column 2/.style={
				column sep=5pt,
			},at={(-0.08,1.01)},anchor=south west,font=\footnotesize}]
		\pgfplotstableread[col sep = ampersand]{sparse.dat}\sparse
		\pgfplotstabletranspose[string type,
		colnames from=rate,
		input colnames to=rate] \sparsetran{\sparse};
		\addplot [color=black,mark=triangle*,mark size=3pt,line width=2pt,]	table [x index ={0}, y index= {8}] {\sparsetran};\addlegendentry{\textbf{AltMinLowRaP(7920)}}
		\addplot [ color=blue,mark=square*,mark size=3pt,line width=2pt,]	table [x index ={0}, y index= {2}] {\sparsetran};\addlegendentry{Block CoPRAM(876)}
		\addplot [ color=olive,mark=square*,mark size=3pt,line width=2pt,]	table [x index ={0}, y index= {3}] {\sparsetran};\addlegendentry{CoPRAM,spatial(1450)}
		\addplot [ color=cyan,mark=square*,mark size=3pt,line width=2pt,]	table [x index ={0}, y index= {4}] {\sparsetran};\addlegendentry{CoPRAM,wavelet(748)}
		\addplot [ color=brown,mark=square*,mark size=3pt,line width=2pt,]	table [x index ={0}, y index= {5}] {\sparsetran};
		\addlegendentry{IERA(2759)}
%
		\end{axis}
		\end{tikzpicture}
}
		\caption{\small{{\bf 
SPMag final files' editor: please help rearrange this figure to look nicer. ?? 
Comparing sparse PR, low rank PR, and unstructured PR solutions for dynamic Fourier ptychographic imaging of the video shown in (a):}
We plot the Structural Similarity Index (SSIM) of the recovered video on the y-axis and undersampling level, $f$, on the x-axis. This is done for five different approaches: IERA (unstructured PR algorithm for Fourier ptychography, applied on each frame); CoPRAM, spatial (sparse PR algorithm applied on each frame individually while assuming the images are sparse; CoPRAM, wavelet (same as CoPRAM spatial, but assumed wavelet sparsity of each image); Block CoPRAM, wavelet (exploits block sparsity across the video sequence);
and AltMinLowRaP (LR PR algorithm along with a model error correction step explained in the text and with using $r=5$). All approaches are implemented with the initialization needed for Fourier ptychography instead of their usual spectral initialization step (see Algorithms 1 and 2 of \cite{TCIgauri}).
Time taken in seconds for $f=$ case is displayed in parentheses next to the algorithm name.
As can be see the low-rank assumption is a much better idea for dynamic imaging in the undersampled (smaller $m$) setting,  but it is also slower.
This figure is taken (re-plotted using data) from \cite{ICIP20}.
}}
\label{bact_comp}
	\end{figure}
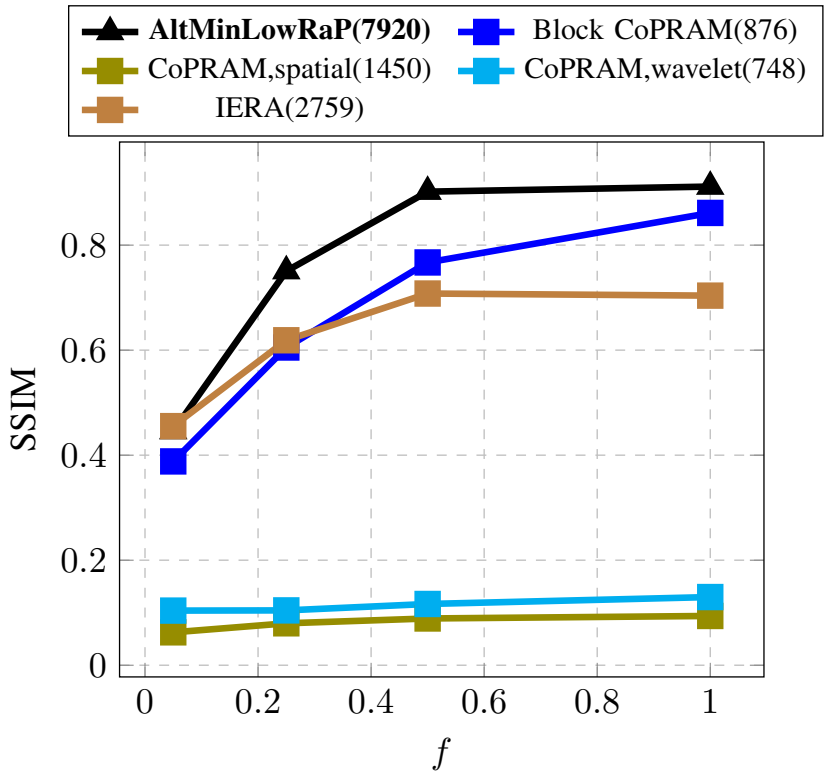

%% file: plot.tex
\begin{tikzpicture}[scale=0.95]
\begin{axis}
[xlabel= Fraction of samples $f$, 
ylabel= SSIM,
grid style = dashed,
grid=both,
legend style=
{at={(1.02,0)},
anchor=south west,
} ,
]

\addplot[line width=2pt,mark=*, mark size=3pt, color=blue] plot coordinates {
	(0,      0)
	(0.2,    0.8928)
	(0.4,    0.8956)
	(0.6,    0.8961)
	(0.8,    0.8990)
	(1.0,    0.9012)
};

\addplot[line width=2pt, mark=*, mark size=3pt, color=orange] plot coordinates {
	(0,     0.0)
	(0.2,    0.8130)
	(0.4,    0.8147)
	(0.6,    0.8153)
	(0.8,    0.8156)
	(1.0,    0.8177)
};

\addplot[color=black, mark=o, line width=2pt, mark size=3pt] plot coordinates {
	(0,     0.0)
	(0.2,    0.7777)
	(0.4,    0.8158)
	(0.6,    0.8412)
	(0.8,    0.8620)
	(1.0,    0.8781)
};

\addplot [color=cyan, mark=o, line width=2pt, mark size=3pt] plot coordinates {
	(0,     0.0)
	(0.2,    0.3358)
	(0.4,    0.3462)
	(0.6,    0.3642)
	(0.8,    0.3948)
	(1.0,    0.4501)
};
\addplot[line width=2pt,mark=*, mark size=3pt, color=red] plot coordinates {
	(0,     0.0)
	(0.2,    0.4135)
	(0.4,    0.4211)
	(0.6,    0.4442)
	(0.8,    0.4548)
	(1.0,    0.5001)
};

\legend{CoPRAM\\Block CoPRAM\\Modified SPARTA\\IERA\\TV regularized\\}

\end{axis}
\end{tikzpicture} 

%% file: StructPR_revise_3.bbl
\begin{thebibliography}{10}

\bibitem{nonconvex_review}
Yuejie Chi, Yue~M Lu, and Yuxin Chen,
\newblock ``Nonconvex optimization meets low-rank matrix factorization: An
  overview,''
\newblock {\em IEEE Transactions on Signal Processing}, vol. 67, no. 20, pp.
  5239--5269, 2019.

\bibitem{ger_saxton}
R.~W. Gerchberg and W.~O. Saxton,
\newblock ``A practical algorithm for the determination of phase from image and
  diffraction plane pictures,''
\newblock {\em Optik}, 1972.

\bibitem{versh_book}
Roman Vershynin,
\newblock {\em High-dimensional probability: An introduction with applications
  in data science}, vol.~47,
\newblock Cambridge University Press, 2018.

\bibitem{holloway}
J.~Holloway, M.~S. Asif, M.~K. Sharma, N.~Matsuda, R.~Horstmeyer, O.~Cossairt,
  and A.~Veeraraghavan,
\newblock ``Toward long-distance subdiffraction imaging using coherent camera
  arrays,''
\newblock {\em IEEE Trans Comput Imaging}, vol. 2, no. 3, pp. 251--265, 2016.

\bibitem{TCIgauri}
G.~Jagatap, Z.~Chen, S.~Nayer, C.~Hegde, and N.~Vaswani,
\newblock ``Sample efficient fourier ptychography for structured data,''
\newblock {\em IEEE Trans Comput Imaging}, 2019.

\bibitem{st_imaging}
Zhi-Pei Liang,
\newblock ``Spatiotemporal imagingwith partially separable functions,''
\newblock in {\em 2007 4th IEEE International Symposium on Biomedical Imaging:
  From Nano to Macro}. IEEE, 2007, pp. 988--991.

\bibitem{dyn_mri1}
Sajan~Goud Lingala, Yue Hu, Edward DiBella, and Mathews Jacob,
\newblock ``Accelerated dynamic mri exploiting sparsity and low-rank structure:
  kt slr,''
\newblock {\em IEEE transactions on medical imaging}, vol. 30, no. 5, pp.
  1042--1054, 2011.

\bibitem{lee2017near}
Kiryung Lee, Yihong Wu, and Yoram Bresler,
\newblock ``Near-optimal compressed sensing of a class of sparse low-rank
  matrices via sparse power factorization,''
\newblock {\em IEEE Transactions on Information Theory}, vol. 64, no. 3, pp.
  1666--1698, 2017.

\bibitem{optspace}
R.~H. Keshavan, A.~Montanari, and S.~Oh,
\newblock ``Matrix completion from a few entries,''
\newblock {\em IEEE Transactions on Information Theory}, vol. 56, no. 6, pp.
  2980--2998, 2010.

\bibitem{lowrank_altmin}
P.~Netrapalli, P.~Jain, and S.~Sanghavi,
\newblock ``Low-rank matrix completion using alternating minimization,''
\newblock in {\em STOC}, 2013.

\bibitem{robpca_nonconvex}
P.~Netrapalli, U~N Niranjan, S.~Sanghavi, A.~Anandkumar, and P.~Jain,
\newblock ``Non-convex robust pca,''
\newblock in {\em NIPS}, 2014.

\bibitem{candes_phaselift}
E.~J. Candes, T.~Strohmer, and V.~Voroninski,
\newblock ``Phaselift: Exact and stable signal recovery from magnitude
  measurements via convex programming,''
\newblock {\em Comm. Pure Appl. Math.}, 2013.

\bibitem{pr_altmin}
P.~Netrapalli, P.~Jain, and S.~Sanghavi,
\newblock ``Phase retrieval using alternating minimization,''
\newblock in {\em NIPS}, 2013, pp. 2796--2804.

\bibitem{altmin_irene_w}
Ir{\`e}ne Waldspurger,
\newblock ``Phase retrieval with random gaussian sensing vectors by alternating
  projections,''
\newblock {\em IEEE Transactions on Information Theory}, vol. 64, no. 5, pp.
  3301--3312, 2018.

\bibitem{wf}
E.~J. Candes, X.~Li, and M.~Soltanolkotabi,
\newblock ``Phase retrieval via wirtinger flow: Theory and algorithms,''
\newblock {\em IEEE Trans. Info. Th.}, vol. 61, no. 4, pp. 1985--2007, 2015.

\bibitem{twf}
Y.~Chen and E.~Candes,
\newblock ``Solving random quadratic systems of equations is nearly as easy as
  solving linear systems,''
\newblock in {\em NIPS}, 2015, pp. 739--747.

\bibitem{lrpr_improved}
S.~Nayer and N.~Vaswani,
\newblock ``Sample-efficient low rank phase retrieval,''
\newblock {\em arXiv:2006.06198}, June 2020.

\bibitem{lrpr_tsp}
N.~Vaswani, S.~Nayer, and Y.~C. Eldar,
\newblock ``Low rank phase retrieval,''
\newblock {\em IEEE Trans. Sig. Proc.}, August 2017.

\bibitem{lrpr_icml}
S.~Nayer, P.~Narayanamurthy, and N.~Vaswani,
\newblock ``Phaseless pca: Low-rank matrix recovery from column-wise phaseless
  measurements,''
\newblock in {\em Intnl. Conf. Machine Learning (ICML)}, 2019.

\bibitem{lrpr_it}
S.~Nayer, P.~Narayanamurthy, and N.~Vaswani,
\newblock ``Provable low rank phase retrieval,''
\newblock {\em IEEE Trans. Info. Th.}, March 2020.

\bibitem{wainwright_linear_columnwise}
Sahand Negahban, Martin~J Wainwright, et~al.,
\newblock ``Estimation of (near) low-rank matrices with noise and
  high-dimensional scaling,''
\newblock {\em The Annals of Statistics}, vol. 39, no. 2, pp. 1069--1097, 2011.

\bibitem{cov_sketch}
Yuxin Chen, Yuejie Chi, and Andrea~J Goldsmith,
\newblock ``Exact and stable covariance estimation from quadratic sampling via
  convex programming,''
\newblock {\em IEEE Transactions on Information Theory}, vol. 61, no. 7, pp.
  4034--4059, 2015.

\bibitem{cai}
T.T. Cai, X.~Li, and Z.~Ma,
\newblock ``Optimal rates of convergence for noisy sparse phase retrieval via
  thresholded wirtinger flow,''
\newblock {\em The Annals of Statistics}, vol. 44, no. 5, pp. 2221--2251, 2016.

\bibitem{voroninski13}
Xiaodong Li and Vladislav Voroninski,
\newblock ``Sparse signal recovery from quadratic measurements via convex
  programming,''
\newblock {\em SIAM Journal on Mathematical Analysis}, vol. 45, no. 5, pp.
  3019--3033, 2013.

\bibitem{jaganathan2013sparse2}
Kishore Jaganathan, Samet Oymak, and Babak Hassibi,
\newblock ``Sparse phase retrieval: Uniqueness guarantees and recovery
  algorithms,''
\newblock {\em IEEE Trans. Sig. Proc.}

\bibitem{sparta}
G.~Wang, L.~Zhang, G.~B. Giannakis, M.~Akcakaya, and J.~Chen.,
\newblock ``Sparse phase retrieval via truncated amplitude flow,''
\newblock {\em arXiv preprint arXiv:1611.07641}, 2016.

\bibitem{fastphase}
G.~Jagatap and C.~Hegde,
\newblock ``Fast sample-efficient algorithms for structured phase retrieval,''
\newblock in {\em Adv. Neural Inf. Proc. Sys. (NIPS)}, Dec. 2017.

\bibitem{ramachandran15}
Ramtin Pedarsani, Dong Yin, Kangwook Lee, and Kannan Ramchandran,
\newblock ``Phasecode: Fast and efficient compressive phase retrieval based on
  sparse-graph codes,''
\newblock {\em IEEE Transactions on Information Theory}, 2017.

\bibitem{bahmani}
S.~Bahmani and J.~Romberg,
\newblock ``Efficient compressive phase retrieval with constrained sensing
  vectors,''
\newblock in {\em Advances in Neural Information Processing Systems}, 2015, pp.
  523--531.

\bibitem{taf}
G.~Wang, G.~B. Giannakis, and Y.~C. Eldar,
\newblock ``Solving systems of random quadratic equations via truncated
  amplitude flow,''
\newblock {\em arXiv preprint arXiv:1605.08285}, 2016.

\bibitem{hughes_icml_2014}
Farhad~Pourkamali Anaraki and Shannon Hughes,
\newblock ``Memory and computation efficient pca via very sparse random
  projections,''
\newblock in {\em International Conference on Machine Learning}, 2014, pp.
  1341--1349.

\bibitem{aarti_singh_subs_learn}
Akshay Krishnamurthy, Martin Azizyan, and Aarti Singh,
\newblock ``Subspace learning from extremely compressed measurements,''
\newblock in {\em Asilomar Conference}, 2014.

\bibitem{lee2019neurips}
Rakshith~Sharma Srinivasa, Kiryung Lee, Marius Junge, and Justin Romberg,
\newblock ``Decentralized sketching of low rank matrices,''
\newblock in {\em Advances in Neural Information Processing Systems}, 2019, pp.
  10101--10110.

\bibitem{rmc_gd}
Y.~Cherapanamjeri, K.~Gupta, and P.~Jain,
\newblock ``Nearly-optimal robust matrix completion,''
\newblock {\em ICML}, 2016.

\bibitem{ICIP20}
Z.~Chen, S.~Nayer, and N.~Vaswani,
\newblock ``Fast and sample-efficient low rank fourier ptychography,''
\newblock in {\em IEEE Intl. Conf. Image Proc. (ICIP)}, submitted, 2020.

\bibitem{pr_generative_prior}
P.~Hand, O.~Leong, and V.~Voroninski,
\newblock ``Phase retrieval under a generative prior,''
\newblock in {\em NIPS}, 2018.

\bibitem{maleki_isit18}
M.~Bakhshizadeh, A.~Maleki, and S.~Jalali,
\newblock ``Compressive phase retrieval of structured signals,''
\newblock in {\em IEEE Intl. Symp. Info. Th. (ISIT)}, 2018.

\bibitem{dyncs_review}
N.~Vaswani and J.~Zhan,
\newblock ``{Recursive Recovery of Sparse Signal Sequences from Compressive
  Measurements: A Review},''
\newblock {\em IEEE Trans. Sig. Proc.}, vol. 64 (13), pp. 3523--3549, 2016.

\bibitem{pr_mc_reuse_meas}
Cong Ma, Kaizheng Wang, Yuejie Chi, and Yuxin Chen,
\newblock ``Implicit regularization in nonconvex statistical estimation:
  Gradient descent converges linearly for phase retrieval, matrix completion
  and blind deconvolution,''
\newblock in {\em Intnl. Conf. Machine Learning (ICML)}, 2018.

\end{thebibliography}
